\pdfoutput=1

\documentclass[11pt]{article}

\usepackage{emnlp2021}

\usepackage{times}
\usepackage{latexsym}
\usepackage{xspace}
\usepackage{amsmath}
\usepackage{caption}
\usepackage{booktabs}
\usepackage{graphicx}
\usepackage{subfigure}
\usepackage[T1]{fontenc}
\usepackage{amsfonts}
\usepackage{paralist}
\usepackage{pgfplots, pgfplotstable}
\usepackage{filecontents}
\usepackage{pgfplots,tikz}
\usepackage{graphicx}
\pgfplotsset{compat=1.14}
\usetikzlibrary{pgfplots.groupplots}
\usepackage{bbm}
\usepackage{enumitem}

\usepackage[utf8]{inputenc}

\usepackage{microtype}

\title{Fairness-aware Class Imbalanced Learning}

\author{Shivashankar Subramanian$^\clubsuit$ \quad Afshin Rahimi$^\spadesuit$ \\{\bf Timothy
  Baldwin$^\clubsuit$ \quad Trevor Cohn$^\clubsuit$ \quad Lea
  Frermann$^\clubsuit$}\\$^\clubsuit$School of Computing and Information
Systems, University
  of Melbourne \\ $^\spadesuit$School of Information Technology and Electrical Engineering, University of Queensland \\
 {\small \url{shivashankarrs@gmail.com} \quad \url{a.rahimi@uq.edu.au}} \\ {\small \url{{tbaldwin, t.cohn, lfrermann}@unimelb.edu.au}}}

\newcommand{\method}[2][]{\textup{#2}\ensuremath{_{\mathrm{#1}}}\xspace}

\newcommand{\vanilla}{\method{vanilla}}
\newcommand{\cw}{\method{CW}}
\newcommand{\iw}{\method{IW}}
\newcommand{\inlp}{\method{INLP}}
\newcommand{\ldam}{\method{LDAM}}
\newcommand{\ldamcw}{\method[cw]{LDAM}}
\newcommand{\ldamreg}{\method[reg]{LDAM}}
\newcommand{\ldamadv}{\method[adv]{LDAM}}
\newcommand{\ldamiw}{\method[iw]{LDAM}}
\newcommand{\focal}{\method{FOCAL}}
\newcommand{\dice}{\method{DICE}}
\newcommand{\dadv}{\method{ADV}}

\newcommand{\ldamL}{\ensuremath{\mathcal{L}_{\ldam}}}
\newcommand{\ldamregL}{\ensuremath{\mathcal{L}_{\ldamreg}}}
\newcommand{\ldamiwL}{\ensuremath{\mathcal{L}_{\ldamiw}}}
\newcommand{\ldamadvL}{\ensuremath{\mathcal{L}_{\ldamadv}}}

\newcommand{\oneGAP}{\ensuremath{1{-}\operatorname{GAP}}\xspace}

\begin{document}

\maketitle

\begin{abstract}
Class imbalance is a common challenge in many NLP tasks, and has clear connections to bias, in that bias in training data often leads to higher accuracy for majority groups at the expense of minority groups. However there has traditionally been a disconnect between research on class-imbalanced learning and mitigating bias, and only recently have the two been looked at through a common lens. In this work we evaluate long-tail learning methods for tweet sentiment and occupation classification, and extend a margin-loss based approach with methods to enforce fairness. We empirically show through controlled experiments that the proposed approaches help mitigate both class imbalance and demographic biases.\footnote{Code available at: \url{https://github.com/shivashankarrs/classimb_fairness}}
\end{abstract}

\section{Introduction}
\label{sec:introduction}
Class imbalance is common in many NLP tasks, including machine reading comprehension \cite{li2020dice}, authorship attribution \cite{caragea2019myth}, toxic language detection \cite{breitfeller-etal-2019-finding}, and text classification \cite{tiangraph}. 
A skewed class distribution hurts the performance of deep learning models \cite{BUDA2018249}, and approaches such as instance weighting \cite{Lin_2017_ICCV, effectivenumber, li2020dice}, data augmentation \cite {juuti2020little, wei-zou-2019-eda}, and weighted max-margin \cite{ldam} are commonly used to alleviate the problem. 

Bias in data often also manifests as skewed distributions, especially when considered in combination with class labels.
This is often referred to as ``stereotyping'' whereby one or more private attributes are associated more frequently with certain target labels, for instance more \textit{men} being employed as \textit{surgeons} than \textit{women}.
Prior work has identified several classes of bias, including bias towards demographic groups based on gender, disability, race or religion \cite{caliskan2017semantics, may2019measuring, garimella2019women, nangia2020crows}, and bias towards individuals \cite{prabhakaran2019perturbation}. Methods to mitigate these biases include data augmentation \cite{www2019}, adversarial learning \cite{li2018towards}, instance weighting based on group membership \cite{Kamiran2011DataPT}, regularization \cite{wickunlocking, kennedy-etal-2020-contextualizing}, and explicit subspace removal \cite{bolukbasi2016man, ravfogel2020null}. 

 \begin{figure}[t]
    \centering
     \includegraphics[scale=0.45]{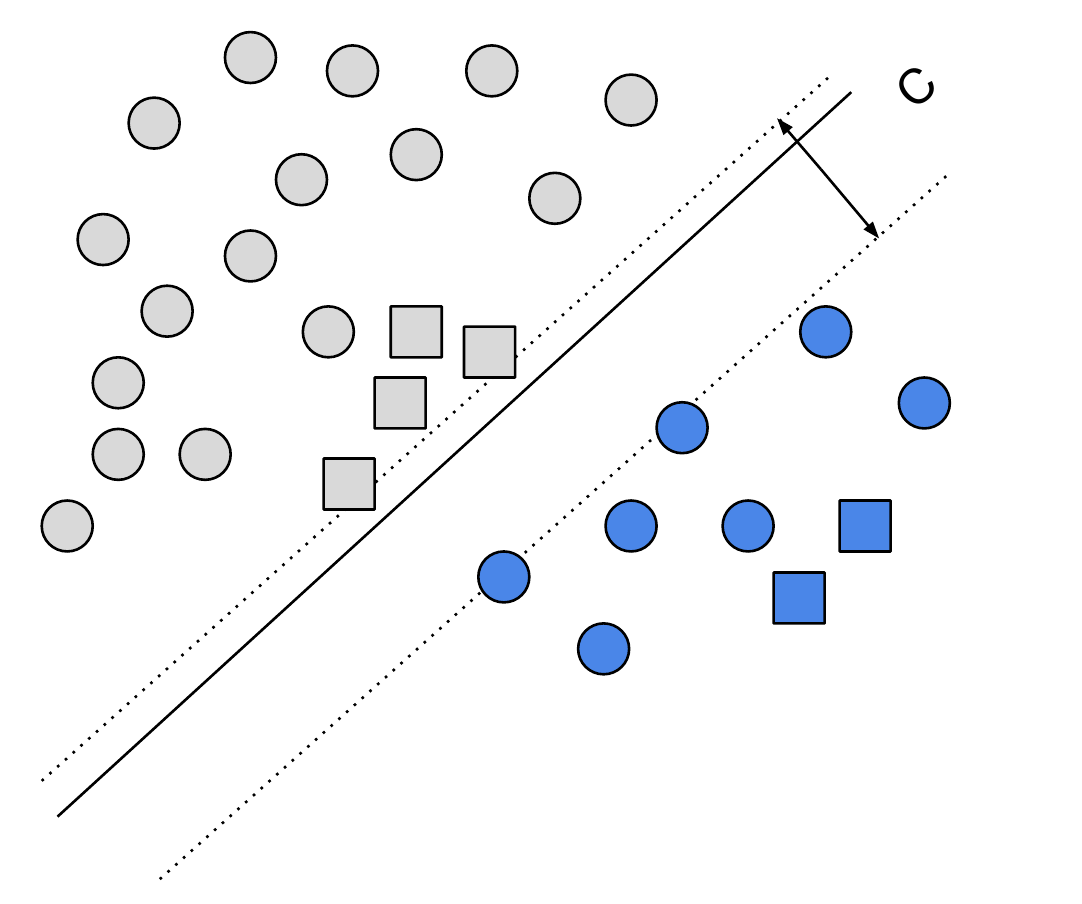}
     \caption{Example of a two-class problem where grey and blue points denote majority and minority classes, respectively, and circles and squares denote two sub-groups. Imbalanced learning methods such as LDAM~\cite{ldam} maximise the (soft-)margin for minority classes and do not consider sub-groups within each class.}
   \label{fig:motivation}
 \end{figure}

This paper draws a connection between class-imbalanced learning and stereotyping bias. Most work has focused on class-imbalanced learning and bias mitigation as separate problems, but the unfairness caused by social biases is often aggravated by the presence of class imbalance \cite{cikm2020}. Class-imbalanced learning approaches improve the performance of minority classes at some cost to the performance of majority classes. A common approach re-weights instances in the training objective to be proportional to the inverse frequency of their class. Approaches such as \focal \citep{Lin_2017_ICCV} and \dice \citep{li2020dice} extend this approach by down-weighting ``easy'' instances. Label-Distribution-Aware Margin Loss (``LDAM'': \citet{ldam}) is an alternative approach, which encourages a larger margin for the minority class, but it does not consider sub-group proportions (see Figure \ref{fig:motivation}). On the other hand, debiasing approaches do not typically focus on class imbalance explicitly.
For instance, in toxicity classification, certain sub-groups are often predicted more confidently for toxicity (encouraging false negatives for the majority sub-group), which tend to be close to the margin for the non-toxic class (encouraging false positives; \citet{10.1145/3308560.3317593}). 

In this work, we modify the training objective of LDAM as a state-of-the-art approach for imbalanced learning so that margins depend not just  on class-imbalance, but also on the subgroup distribution within each class. Specifically, we extend LDAM with popular debiasing strategies. We show the effectiveness of our approach through several controlled experiments on two text classification data sets.

\section{Proposed Approaches}
\label{sec:approach}


Let $(\mathbf{x},y,g)$ denote a training instance, comprising an input, label, and group identifier, respectively.
\ldam~\cite{ldam} addresses class imbalance by enforcing a larger margin for minority classes:
\begin{align*}
\ldamL(\mathbf{x},y;f) & = -\log \frac{e^{z_{y}-\Delta_{y}}}{e^{z_{y}-\Delta_{y}} + \sum_{j \neq y} e^{z_j}} \\
  \Delta_j &= \frac{C}{n_{j}^{1/4}} \textup{ for } j \in \{1,\dots, k\} 
\end{align*}
where $\mathbf{z} = f(\mathbf{x})$ are the model outputs, $k$ is the number of classes, $n_j$ is the number of instances in class $j$, and $C$ is a hyperparameter.
Smaller classes are associated with a larger $\Delta_{y}$, which is subtracted from the model output $z_y$, thus enforcing a larger margin. 

We propose three extensions to \ldam, each of which takes into account imbalance in the distribution of private attributes across classes:\
\paragraph{\ldamiw} adds instance re-weighting, based on groups within each class:
\begin{align*}
  \ldamiwL(\mathbf{x},y,g;f) = \mathrm{\omega}_{y, g} \ldamL(x,y;f) 
\end{align*}
where $g$ is the group of instance $\mathbf{x}$; and $\mathrm{\omega}_{y,g}=\frac{1-\beta}{1-\beta^{N_{y,g}}}$
weights each class--group combination based on its smoothed inverse frequency. $\beta$ is a constant set to 0.9999 \cite{effectivenumber} and $N_{y, g}$ is the number of instances belonging to class $y$ and group $g$.
Mistakes on minority groups within minority classes are penalised most.

\paragraph{\ldamadv} adds an adversarial term, so that the learnt representations are a poor predictor of group membership \cite{li2018towards}:
\begin{align*}
 \ldamadvL&(\mathbf{x},y,g;f) = \nonumber \\
 & \ldamL(\mathbf{x},y;f) -  \lambda_{\text{adv}}  \mathrm{CE}(g, l(\mathbf{x})) 
\end{align*}
where $f$ shares the lower layers of the network with the adversary $l$,  $\mathrm{CE}$ denotes cross-entropy loss, and $\lambda_\text{adv}$ is a hyperparameter. 
This objective $\ldamadvL$ is jointly minimised \emph{wrt}  $f$ and maximised \emph{wrt} $l$. The penalty results in hidden representations that are informative for the main classification task ($f$), but uninformative for the adversarial group membership prediction task ($l$). The adversarial loss is implemented using gradient reversal~\cite{gradientreversal}.


\paragraph{\ldamreg} 
adds a soft regularization term which encourages fairness as \textit{equalised odds} by reducing maximum mean discrepancy \cite{gretton2012kernel} across groups. The probability of predicting some class $k$ for any individual group $g$ should be close to $k$'s probability over the whole data set:
\begin{align*}
  &\ldamregL(\mathbf{X},\mathbf{y},\mathbf{g};f) = \ldamL(\mathbf{X},\mathbf{y};f) + \nonumber \\ 
  &\quad  \rho \sum_g \left\|  \frac{1}{N_g} \sum_{i: g_i=g} f(\mathbf{x}_i) - \frac{1}{N} \sum_i f(\mathbf{x}_i)\right\|^2
    \label{eq:reg}
\end{align*}
where we have moved from single instance loss to the loss over the full training set, 
$N_g$ denotes the number of training instances in group $g$,
and hyper-parameter $\rho$ controls the trade-off between performance and fairness. 

\section{Experimental Results}

We perform experiments on the two tasks of emoji prediction and occupation classification, both of which are binary classification tasks with binary protected attributes.

\begin{description}
\item{\textbf{Emoji prediction:}} We use the Twitter dataset of  \citet{blodgett-etal-2016-demographic}, where tweets are associated with the private attribute race ({black}/{white}), and sentiment labels are derived from emoji usage ({happy}/{sad}) \cite{elazar-goldberg-2018-adversarial}. We experiment with different levels of class and stereotyping imbalance in the Emoji dataset, including its original distribution (see Section \ref{sec:mc}). 

\item{\textbf{Occupation classification:}} This data set consists of short biographies scraped from the web, annotated for private attribute gender ({male}/{female}) and target occupation labels \cite{10.1145/3287560.3287572}. We focus on two occupations with well-documented gender stereotypes -- {surgeon} and {nurse}. The resulting dataset is mildly class-imbalanced (59\% surgeon: 41\% {nurse}), with roughly symmetric natural gender splits (90\% male for {surgeon} and 90\% female for {nurse}).
\end{description}
For emoji prediction we follow \citet{ravfogel2020null} and use the DeepMoji encoder, which was trained on millions of tweets and is known to encode demographic information~\cite{elazar-goldberg-2018-adversarial}. For occupation classification, we use the BERT-base uncased model and classify via the last hidden state of the CLS token \cite{devlin2019bert}. Both encoders are followed by a single hidden layer MLP. 

We evaluate classification performance based on macro-averaged F-score (to account for class imbalance), and evaluate fairness using performance GAP: the average of the true positive rate (TPR) and true negative rate (TNR) differences between the two subgroups \cite{10.1145/3287560.3287572, ravfogel2020null}. Note that a wide variety of fairness measures (both on the group- and individual levels) have been proposed, which are impossible to satisfy simultaneously
\cite{garg2020fairness}. Often, a suitable measure is chosen based on the target application. Here we use the popular \textit{equalised odds} measure considering both TPR and TNR of classifiers, in order to address scenarios where certain sub-groups are predicted more often with some classes (see Section \ref{sec:introduction}).  We report fairness as $1-$GAP, such that higher numbers are better, and a perfectly fair model achieves $\oneGAP = 1$. We compare our methods against the following benchmarks:
\begin{description}
\item[\vanilla:] unweighted cross-entropy loss.
\item[\focal:] re-weights easy examples during training \cite{Lin_2017_ICCV}.
\item[\cw:] instance re-weighting based on the inverse class proportion and cross-entropy.
\item[\iw:] instance re-weighting based on the combination of inverse class and group proportions, and cross-entropy \cite{Kamiran2011DataPT}. 
\item[\inlp:] Iterative null-space projection \cite{ravfogel2020null}: in each iteration, we learn a SVM classifier $W$ using hidden representations ($X_h$) as the independent variables to predict the protected attribute, where $X_h$ is projected onto the nullspace of $W$ 
to remove the protected information.
\item[\ldam:] the original \ldam model~\cite{ldam}. 
\item[\ldamcw:] a variant of \ldam with instance re-weighting by inverse class proportion \cite{ldam}.
\end{description}

\newcommand{\plotfile}[1]{
	\pgfplotstableread{#1}{\table}
	\pgfplotstablegetcolsof{#1}
	\pgfmathtruncatemacro\numberofcols{\pgfplotsretval-1}
	\pgfplotsinvokeforeach{1,...,\numberofcols}{
		\pgfplotstablegetcolumnnamebyindex{##1}\of{\table}\to{\colname}
			\addplot+[line width=2pt,mark=none] table [y index=##1] {#1};
	}
}

\pgfplotsset{
	axis background/.style={fill=white},
	ytick pos=left,
	tick style={
		major grid style={style=white,line width=1pt},
		tick align=outside,
	},
	commonstyle/.style={
		draw=white,
		mark=*,
	},
	midystyle/.style = {
		yticklabels={,,},
		ytick style={draw=none},
		ylabel = {},
	},
	midystyle2/.style = {
	ytick style={},
	tick align =outside,
	ylabel = {},
	},
	midxstyle/.style = {
		xtick style={draw=none},
		xlabel = {},
	},
	cossimstyle/.style = {
		ymin = 0.8,
	},
}

\definecolor{mygrey}{RGB}{229,229,229}
\definecolor{mygrey2}{RGB}{127,127,127}
\definecolor{mygrey3}{RGB}{240,240,240}
\definecolor{cLarge}{RGB}{31,120,180}
\definecolor{cVerySmall}{RGB}{253,191,111}
\definecolor{mycolor1}{RGB}{166,206,227}
\definecolor{mycolor2}{RGB}{31,120,180}
\definecolor{mycolor3}{RGB}{178,223,138}
\definecolor{mycolor4}{RGB}{51,160,44}
\definecolor{mycolor5}{RGB}{251,154,153}
\definecolor{mycolor6}{RGB}{227,26,28}
\definecolor{mycolor7}{RGB}{253,191,111}
\definecolor{mycolor8}{RGB}{255,127,0}
\definecolor{mycolor9}{RGB}{0, 0, 0}
\definecolor{mycolor10}{RGB}{106,61,154}

\usepgfplotslibrary{colorbrewer}
\pgfplotsset{
	compat=1.13,
	my style/.style={
		ylabel=1 - GAP,
		axis lines*       = left,
		every axis title/.append style={at={(0.5,-0.5)}},
	},
	cycle list/Dark2,
	my legend style/.style={
		legend entries={
		\ldamreg,
		\ldamadv,
		\inlp,
		\ldam,
		\ldamcw,
		\ldamiw,
		\focal,
		\vanilla,
		\iw,
		\cw,
		},
		legend style={
 			at={([yshift=2pt]0,1)},
			anchor=south west,
			align=center,
			fill=none,
			draw=white,
                        font=\small,
		},
		legend columns=10,
	},
		points/.style={only marks, mark size=4},
		cycle multiindex* list={
			mycolor7,
			mycolor2,
			mycolor3,
			mycolor4,
			mycolor5,
			mycolor6,
			mycolor1,
			mycolor8,
			mycolor9,
			mycolor10
			\nextlist
			mark=x,
			mark=star,
			mark=halfcircle,
			mark=diamond*,
			mark=otimes*,
			mark=square*,
			mark=oplus*,
			mark=heart,
			mark=triangle*,
			mark=pentagon*
			\nextlist
		},
}

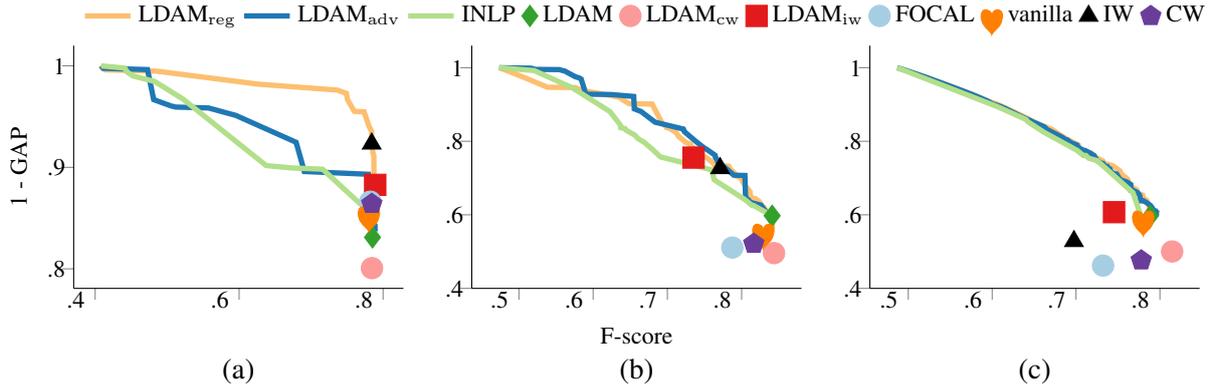
\begin{figure*}[t]
	\centering

\begin{tikzpicture}%
\begin{groupplot}[%
group style={%
	group name=QuantileError,%
	group size= 3 by 1,%
	horizontal sep = 0.9cm, 
},
my style,%
width=0.37\textwidth,
height=0.3\textwidth,
legend cell align={left}, %
every axis label/.style={font=\small},%
ticklabel style = {font=\small},%
yticklabel style={xshift=5pt,/pgf/number                           
		format/.cd,fixed,precision=1,/tikz/.cd},%
x tick label style={rotate=0, anchor=east},
xticklabel style={/pgf/number                           
	format/.cd,fixed,precision=1,zerofill,/tikz/.cd},
]%

\nextgroupplot[%
my legend style,
title={(a)},
ytick={0.8, 0.9, 1},
yticklabels={.8, .9, 1},
xtick={0.4,0.6,0.8},
xticklabels={.4,.6,.8},
]%
\plotfile{ldamreg1.dat}
\plotfile{ldamadv1.dat}
\plotfile{inlp1.dat}
\addplot+[points] coordinates {(0.785272745,0.83101358)};
\addplot+[points] coordinates {(0.784570373,0.800678784)};
\addplot+[points] coordinates {(0.789060502,0.882774428)};
\addplot+[points] coordinates {(0.782198995,0.86614954)};
\addplot+[points] coordinates {(0.780276054,0.85506849)};
\addplot+[points] coordinates {(0.784168265,0.923269217)};
\addplot+[points] coordinates {(0.784336028,0.864112698)};	
\nextgroupplot[%
midystyle2,
title={(b)},
xlabel=F-score,
ymin=0.4,
ytick={0.4, 0.6,0.8,1},
yticklabels={.4, .6,.8,1},
xtick={0.5,0.6,0.7,0.8},
xticklabels={.5,.6,.7,.8},
]%
\plotfile{ldamreg2.dat}
\plotfile{ldamadv2.dat}
\plotfile{inlp2.dat}
	\addplot+[points] coordinates {(0.841091771,0.597455164)};
	\addplot+[points] coordinates {(0.843614124,0.495880701)};
	\addplot+[points] coordinates {(0.735282575,0.755602999)};
	\addplot+[points] coordinates {(0.78747503,0.511006994)};
	\addplot+[points] coordinates {(0.829144767,0.557687199)};
	\addplot+[points] coordinates {(0.771032103,0.726499063)};
	\addplot+[points] coordinates {(0.816333704,0.521572357)};
\nextgroupplot[%
midystyle2,
title={(c)},
ymin=0.4,
ytick={0.4, 0.6,0.8,1},
yticklabels={.4, .6,.8,1},
xtick={0.5,0.6,0.7,0.8},
xticklabels={.5,.6,.7,.8},
]%
\plotfile{ldamreg3.dat}
\plotfile{ldamadv3.dat}
\plotfile{inlp3.dat}
\addplot+[points] coordinates {(0.788433692,0.596618477)};
\addplot+[points] coordinates {(0.814571005,0.50034577)};
\addplot+[points] coordinates {(0.74534418,0.60716444)};
\addplot+[points] coordinates {(0.732206406,0.461913892)};
\addplot+[points] coordinates {(0.780083319,0.593356889)};
\addplot+[points] coordinates {(0.697537884,0.528161925)};
\addplot+[points] coordinates {(0.777745634,0.476314245)};
\end{groupplot}
\end{tikzpicture}
\caption{F-score vs.\ fairness (\oneGAP) across the (a) original, (b) 90/90 and (c) 95/95 setting on the emoji prediction task. Models which balance fairness vs.\ performance through hyperparameters are shown as Pareto frontiers, while the others are reported as single points.}
\label{fig:fig123}
\end{figure*}


\pgfplotsset{
	axis background/.style={fill=white},
	ytick pos=left,
	tick style={
		major grid style={style=white,line width=1pt},
		tick align=outside,
	},
	commonstyle/.style={
		draw=white,
		mark=*,
	},
	midystyle/.style = {
		yticklabels={,,},
		ytick style={draw=none},
		ylabel = {},
	},
	midystyle2/.style = {
		ytick style={},
		tick align =outside,
		ylabel = {},
	},
	midxstyle/.style = {
		xtick style={draw=none},
		xlabel = {},
	},
	cossimstyle/.style = {
		ymin = 0.8,
	},
}

\definecolor{mygrey}{RGB}{229,229,229}
\definecolor{mygrey2}{RGB}{127,127,127}
\definecolor{mygrey3}{RGB}{240,240,240}
\definecolor{cLarge}{RGB}{31,120,180}
\definecolor{cVerySmall}{RGB}{253,191,111}
\definecolor{mycolor1}{RGB}{166,206,227}
\definecolor{mycolor2}{RGB}{31,120,180}
\definecolor{mycolor3}{RGB}{178,223,138}
\definecolor{mycolor4}{RGB}{51,160,44}
\definecolor{mycolor5}{RGB}{251,154,153}
\definecolor{mycolor6}{RGB}{227,26,28}
\definecolor{mycolor7}{RGB}{253,191,111}
\definecolor{mycolor8}{RGB}{255,127,0}
\definecolor{mycolor9}{RGB}{0, 0, 0}
\definecolor{mycolor10}{RGB}{106,61,154}

\usepgfplotslibrary{colorbrewer}
\pgfplotsset{
	compat=1.13,
	my style/.style={
		ylabel=1 - GAP,
		xlabel=F-score,
		axis lines*       = left,
		every axis title/.append style={at={(0.3,0.05)}},
	},
	cycle list/Dark2,
	my legend style/.style={
		legend entries={
			\ldamreg,
			\ldamadv,
			\inlp,
			\ldam,
			\ldamcw,
			\ldamiw,
			\focal,
			\vanilla,
			\iw,
			\cw,
		},
		legend style={
			at={([yshift=0.5pt, xshift=0.5pt]0,0)},
			anchor=south west,
			align=center,
			fill=none,
			draw=white,
			font=\tiny,
		},
		legend columns=2,
	},
	points/.style={only marks, mark size=3},
	cycle multiindex* list={
		mycolor7,
		mycolor2,
		mycolor3,
		mycolor4,
		mycolor5,
		mycolor6,
		mycolor1,
		mycolor8,
		mycolor9,
		mycolor10
		\nextlist
		mark=x,
		mark=star,
		mark=halfcircle,
		mark=diamond*,
		mark=otimes*,
		mark=square*,
		mark=oplus*,
		mark=heart,
		mark=triangle*,
		mark=pentagon*
		\nextlist
	},
}

\begin{figure}[t]
	\centering
	\begin{tikzpicture}%
	\begin{axis}[
		my style,%
		width=0.45\textwidth,
		height=0.3\textwidth,
		legend cell align={left}, %
		every axis label/.style={font=\small},%
		ytick style={draw=none},
		xtick style={draw=none},
		ticklabel style = {font=\small},%
		yticklabel style={xshift=5pt,/pgf/number                           
			format/.cd,fixed,precision=2,/tikz/.cd},%
		x tick label style={rotate=0, anchor=east},
		xticklabel style={/pgf/number                           
			format/.cd,fixed,precision=1,zerofill,/tikz/.cd},
		my legend style,
		ytick={0.8, 0.9, 1},
		yticklabels={.8, .9, 1},
		xtick={0.4,0.6,0.8,1},
		xticklabels={.4,.6,.8,1},
	]%
	\plotfile{ldamreg4.dat}
	\plotfile{ldamadv4.dat}
	\plotfile{inlp4.dat}
	\addplot+[points] coordinates {(0.949023332,0.839729629)};
	\addplot+[points] coordinates {(0.949982441,0.841888246)};
	\addplot+[points] coordinates {(0.932398164,0.953371097)};
	\addplot+[points] coordinates {(0.95466279,0.850291789)};
	\addplot+[points] coordinates {(0.953342294,0.86013177)};
	\addplot+[points] coordinates {(0.95212171,0.879483265)};
	\addplot+[points] coordinates {(0.953746743,0.857372422)};	
	\end{axis}
	\end{tikzpicture}
	\caption{F-score vs.\ \oneGAP on occupation classification (original class balance and stereotyping) for the same set of models as in Figure~\ref{tab:inlp_table1}.}
	\label{fig:bioscontrol}
\end{figure}
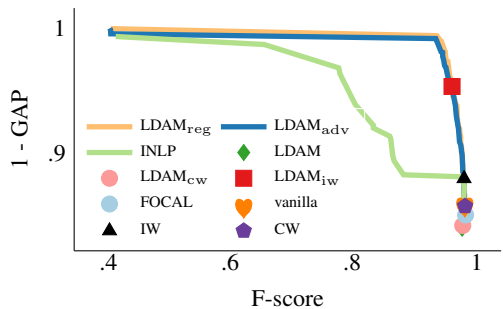

\subsection{Model Comparison}
\label{sec:mc}


  

  

\begin{table*}[!t]
    \centering
    \scalebox{0.75}{
    \begin{tabular}{lcccccccccccccc}
    \toprule
               &  \multicolumn{6}{c}{F-score}  &&   \multicolumn{6}{c}{\oneGAP}   \\
      \cmidrule{2-7}
      \cmidrule{9-14}
    Ratio & \vanilla & \inlp & \ldamiw & \ldamreg & \ldamadv& \dadv&& \vanilla & \inlp & \ldamiw& \ldamreg & \ldamadv & \dadv \\
    \midrule
        0.5  &  0.76     & 0.75  & 0.76 & 0.75  &  0.75 & 0.76      && 0.87  & 0.88  & 0.89 & \bf{0.92} & 0.91 &   0.83    \\
        0.6  &  0.76     & 0.71  & 0.76 & 0.74  &  0.74 & 0.75      && 0.79  & 0.82  & 0.80 & \bf{0.92} & 0.91 & 0.80\\
        0.7  &    0.74   & 0.65  & 0.75 & 0.74  &  0.73 & 0.75      && 0.70  & 0.84  &0.73  & \bf{0.93} & 0.89 &  0.78     \\
        0.8 &    0.72    & 0.62  & 0.74 & 0.73  &  0.73 & 0.74      && 0.61  & 0.84  & 0.67 & \bf{0.93} & 0.72 &  0.76  \\
        \bottomrule
    \end{tabular}}
  \caption{Performance and Fairness on the Emoji data set with fixed balanced class-distribution, but varying the stereotyping ratio (\textit{black}:\textit{white}) per class. The ratio column denotes the \% of \textit{black} instances relative to \textit{white}. The test-set is stereotype balanced (50:50).}
    \label{tab:inlp_table1}
\end{table*}

\begin{table*}[!t]
    \centering
    \scalebox{0.71}{
    \begin{tabular}{ccccccccccccccc}
    \toprule
               &  \multicolumn{6}{c}{F-score}  &&   \multicolumn{6}{c}{\oneGAP}   \\
      \cmidrule{2-7}
      \cmidrule{9-14}
    Ratio & \vanilla & \inlp & \ldamcw & \ldamiw & \ldamreg & \ldamadv&& \vanilla & \inlp & \ldamcw & \ldamiw& \ldamreg & \ldamadv \\
    \midrule
        0.7  & 0.83    & 0.80 &0.83& 0.82 & 0.80 &  0.80   && 0.62  & 0.73 & 0.60& 0.75 & \bf{0.84} & 0.83\\
        0.8  & 0.80    & 0.77 &0.83& 0.80 & 0.77 &  0.78   &&  0.64 & 0.80 & 0.61& 0.76 & 0.84 & \bf{0.85} \\
        0.9 &  0.74   & 0.72 &0.79 & 0.75 & 0.72 &  0.74   &&  0.70 & 0.79 &0.62 & 0.84 & \bf{0.85} & 0.82 \\
        \bottomrule
    \end{tabular}}
  \caption{Performance and Fairness on the Emoji data set, fixing the stereotyping ratio to 0.8:0.2 (\textit{black}:\textit{white}) per class, and varying the class-balance ratio (proportion of positive class is shown). The test sets in each row are different, and mimic the class-imbalance and stereotyping of the training data (i.e.\ results across rows are not comparable).} 
    \label{tab:inlp_table2}
\end{table*}

We include simulated experimental settings with the emoji dataset following \citet{ravfogel2020null} where they keep the class proportions balanced, but vary group proportions (stereotyping). 
In our work, we systematically vary both class imbalance and stereotyping, in order to assess the robustness of the models wrt class imbalance and fairness individually. We explore three settings: varying both dimensions at the same time (Figure \ref{fig:fig123}), controlling for class imbalance and vary stereotyping (Table \ref{tab:inlp_table1}), and controlling for stereotyping while varying class imbalance (Table \ref{tab:inlp_table2}).

We simultaneously vary stereotyping and class imbalance in the emoji dataset, exploring several settings:
\begin{itemize}
\item Original: the dataset is sampled based on the natural class distribution (70\% positive; \newcite{blodgett-etal-2016-demographic}), and within each class the black:white ratio is set to 18:82, based on US census estimates.
\item 90/90: the class distribution is skewed (90\% positive), and black:white ratio is set to 90:10 for positive tweets and 10:90 for negative tweets (i.e.\ ``stereotyping'' the classes).
\item 95/95: as per the above, but with class skew and stereotyping ratios set to 95:5.
\end{itemize}
For the occupation classification task, the original data is used as is (Figure \ref{fig:bioscontrol}).

\paragraph{Model Selection.} For models with hyper-parameters which trade off performance and fairness, the optimal balance of F-score and fairness is not clear, so we adopt the concept of Pareto optimality \cite{godfrey2007algorithms} and present the Pareto frontier in the graphs. In particular, for \ldamreg and \ldamadv, we perform a hyper-parameter search over $C$ ($10^{-2}$ to 30), $\rho$ ($10^{-4}$ to $10^{2}$), and $\lambda$ ($10^{-4}$ to $10^{2}$). In general, a higher $C$ prioritises F-score over fairness, and a higher $\rho$ and $\lambda$ prioritise fairness. For INLP, we tune the number of iterations as a hyper-parameter. The remaining models don't have trade-off hyper-parameters, so we report a single-point best model: for \ldam and \ldamcw we tune $C$ by choosing the best-performing model over the dev set. For \ldamiw, we set $\beta$ to 0.9999 following~\citet{effectivenumber}, and tune $C$ to identify the fairest model on the dev set. 


The results in Figure~\ref{fig:fig123} (a)--(c) show that \ldamreg is overall superior to the other approaches, especially for higher F-scores. For increasingly extreme levels of class imbalance and stereotyping (as we move to the right in the figure), the advantage of \ldamreg over \ldamadv and \inlp decreases substantially. Across all the settings, \ldamcw has the highest bias (is least fair). With higher class imbalance and stereotyping, most class-imbalanced learning methods---\focal, \cw and \ldamcw---exhibit high bias. In the stereotyping settings, \ldamiw reduces bias compared to \iw.

Analogous results on the occupation data (original class proportions) are in Figure \ref{fig:bioscontrol}. Once again, the proposed \ldam extensions perform the best overall, with \ldamreg achieving the best trade-off in performance. Class-imbalanced learning approaches (\focal, \ldam, and \ldamcw) are most biased on this dataset, with \iw improving fairness over \vanilla cross-entropy training, and \ldamiw providing large improvements in fairness.

\subsection{Stereotyping-Class balance Trade-off}

In addition to comparing models based on the trade-off between performance and fairness in class-imbalanced learning, we wish to disentangle the effect of stereotyping from class imbalance. We do so by: (a) fixing class balance to 50:50 and varying stereotyping (Table~\ref{tab:inlp_table1}); and (b) fixing stereotyping to a symmetric 0.8 while varying class imbalance (Table~\ref{tab:inlp_table2}). A stereotyping level of symmetric 0.8 means 80:20 black:white for positive and 20:80 black:white for negative tweets. We perform model selection by choosing the \inlp model with best harmonic mean of performance and fairness for Table~\ref{tab:inlp_table2}, and use the results from the original paper for Table~\ref{tab:inlp_table1}. We select all other models by first selecting from models with F-score at least as high as \inlp, and then selecting the one with the lowest GAP. We include a recent adversarial model in the varying stereotyping experiments, which performed strongly on the class-balanced emoji data (\dadv: \citet{han2021diverse}).

Our results on varying stereotyping levels in Table~\ref{tab:inlp_table1} show that the \vanilla baseline drops in performance more sharply than most proposed models, and results in the most unfair predictions by a large margin. \ldamiw, \ldamadv, and \dadv retain high F-scores but drop in fairness with increasing stereotyping, while \inlp exhibits the opposite pattern.  \ldamreg achieves the best balance of F-score and fairness. Table~\ref{tab:inlp_table2} presents results for fixed stereotyping and varying class imbalance (0.7--0.9 positive). We include \ldamcw for handling class imbalance but exclude \dadv, which does not address class-imbalance directly. We observe that \ldamcw has the highest F-score, but scores poorly for fairness. \ldamiw achieves the best trade-off with high class-imbalance, but shows large variation across settings. \ldamreg appears more stable, exhibiting a good performance--fairness trade-off.


\section{Conclusion and Future Work}

We explored the interplay of class-imbalance and stereotyping in two language classification data sets. We showed that vanilla class-imbalanced learning (\iw, \cw, \focal, \ldam and \ldamcw) can exacerbate unfairness. We extended class-imbalanced learning approaches to handle fairness under stereotyping, and showed that our models provide consistent gains in fairness without sacrificing accuracy. Both \ldamreg which uses maximum mean discrepancy regularizer \cite{tzeng2014deep} and \ldamadv with adversarial loss \cite{gradientreversal} are different ways to make the text representation independent of demographic attributes. Consistent with previous work \cite{louizos2016variational} we find that \ldamreg is robust and performs best across several test scenarios, except in extremely skewed (or stereotyped) settings where the gains of \ldamreg over its adversarial counterpart (\ldamadv) diminishes. In addition, \ldamadv introduces more parameters into the model, and is in general hard to train, hence \ldamreg is more preferable overall. In the future, we plan to extend our methods to more complex tasks and  multiple private attributes \cite{shiva-emnlp-2021-gerry}.




\section{Acknowledgement}
We thank Xudong Han for the discussions and inputs. This  work  was  funded  in  part by the Australian Government Research Training Program Scholarship, and the Australian Research Council.

\bibliography{anthology,custom}
\bibliographystyle{acl_natbib}

\appendix

\end{document}